# Federated Active Learning Framework for Efficient Annotation Strategy in Skin-lesion Classification

Zhipeng Deng*, Yuqiao Yang, Kenji Suzuki

Biomedical Artificial Intelligence Research Unit (BMAI), Institute of Innovative Research, Department of Information and Communications Engineering, School of Engineering, Tokyo Institute of Technology

*Abstract—*

**Federated Learning (FL) enables multiple institutes to train models collaboratively without sharing private data. Current FL research focuses on communication efficiency, privacy protection, and personalization and assumes that the data of FL have already been ideally collected. In medical scenarios, however, data annotation demands both expertise and intensive labor, which is a critical problem in FL. Active learning (AL), has shown promising performance in reducing the number of data annotations in medical image analysis. We propose a federated AL (FedAL) framework in which AL is executed periodically and interactively under FL. We exploit a local model in each hospital and a global model acquired from FL to construct an ensemble. We use ensemble-entropy-based AL as an efficient data-annotation strategy in FL. Therefore, our FedAL framework can decrease the amount of annotated data and preserve patient privacy while maintaining the performance of FL. To our knowledge, this is the first FedAL framework applied to medical images. We validated our framework on real-world dermoscopic datasets. Using only 50% of samples, our framework was able to achieve state-of-the-art performance on a skin-lesion classification task. Our framework performed better than several state-of-the-art AL methods under FL and achieved comparable performance to full-data FL.**

## INTRODUCTION

Deep learning (DL) has become a de facto solution in various fields (LeCun et al.), including computer-aided diagnosis (CAD) (Cheng et al., Jin et al., Suzuki, Suzuki et al., Suzuki et al.). DL can automatically learn complex features directly from the images without the need for feature engineering. However, as a data-driven method, DL's further success in CAD faces two key challenges: 1) lack of annotated data (Budd et al.) and 2) protection of patient privacy (Kaissis et al.). DL methods generally rely on a massive amount of data with accurate annotations. While the vast amount of data available today fuel numerous technological advancements, obtaining such extensive databases in the medical field can prove to be challenging or even unattainable, particularly for less common diseases. Protecting patient privacy is also becoming a public concern in this big-data era. More governments or organizations have established laws and regulations, such as *General Data Protection Regulation* (GDPR) proposed by the European Union, to limit illegal access to private data. In the United States, a national law regarding medical privacy is the *Health Insurance Portability*

---

* Correspondence to: deng.z.aa@m.titech.ac.jp

*and Accountability* (HIPAA) (Act), which aims to enhance privacy rights and protections within medical institutions.

Federated learning (FL) (McMahan et al.) is a decentralized learning paradigm that trains a model across multiple decentralized edge clients without exchanging data. The term *federated learning* was initially introduced by the Google group to solve the issue of heavy data-transfer communication costs and computing power consumption on mobile and edge device applications (McMahan et al.). However, FL was soon applied in other fields where only a small number of clients exists to overcome the privacy-protection and data-limitation problems, which enables multiple nodes to train a model collaboratively without sharing data.–FL has been introduced to the medical field (Chen et al., Rieke et al.) to encourage cross-institutional collaboration considering FL's privacy-preserving property. FL enables hospitals or institutes to train models with their data locally within the hospitals/institutions and send the models to a central server instead of patient data. The central server aggregates these models globally without accessing any patient data in those hospitals or institutes that participated in training the DL model (McMahan et al.). In a recent study for the artificial intelligence-driven screening test for COVID-19 (Soltan et al., 2024), they have shown that the federated training of a global deep neural network classifier improved the performance of the locally trained models by 27.6% in AUROC. Therefore, hospitals participating in FL can build a more generalized and powerful model while ruling out the possibility of leaking patient privacy.

To reduce the workload of data annotation in CAD, integrated active learning (AL) was introduced with DL under centralized training (Shi et al., Zhou et al.). AL methods improve the efficiency of data annotation by iteratively selecting more valuable and informative samples for experts to annotate. With the strategy of eliminating the number of training samples by selecting cases that have the potential to make contributions to the model, AL methods can decrease the number of training samples while maintaining performance. AL has been extensively studied as an annotation strategy in medical-image analysis (Budd et al., Shi et al., Zhou et al.) and proven to be able to achieve state-of-the-art performance with less training data for several medical-image-analysis tasks.

Despite AL's success in centralized learning, with almost all FL (i.e., decentralized learning) methods it is assumed that the FL data have already been ideally collected without embedding the data annotation strategy. Nevertheless, the data-annotation step in FL should not be ignored, which requires a heavy workload and expertise of clinicians to conduct annotation. Therefore, we propose a federated AL (FedAL) framework to meet the need of a data-annotation strategy in FL. Due to the limited number of annotated clinical cases of individual hospitals or institutes, such organizations usually conduct AL in a centralized learning manner, where selected data need to be progressively added into a centralized dataset. This strategy may be difficult to conduct due to the privacy restriction of sharing patient data in each institution or hospital. Accordingly, centralized learning in medical-image analysis by collecting data into a centralized dataset is contradictory to AL to some extent. In contrast, we developed our FedAL framework to encourage hospitals and institutes to collect data in a progressive manner and without leaking privacy, which might indicate that AL is more congruent with FL than using AL for centralized learning in medical-image analysis.

We developed our FedAL framework to improve annotation efficiency in FL with limited training data on clients. Considering the characteristics of FL, we seek a proper means to embed AL into FL. Specifically, our framework uses a global aggregated model and local models to calculate ensemble entropy (Beluch et al.), which is used as the criterion in the data-selection step. The key contributions of our work are as follows.

- We propose a FedAL framework, which meets the need of a data-annotation strategy in FL. The annotation process is simulated using a fully annotated dataset.

- We introduce the novel AL strategy of our framework that can select data valuable for both local models and global models. Therefore, our framework is robust for a not independent and identically distributed (non-IID) FL dataset.

- We compared various methods in a real-world skin-lesion-classification task. These frameworks are 1) FL with random sampling, 2) FL with client-level AL, 3) a current FedAL framework (Ahn et al.) claimed to be state-of-the-art in natural-image-classification tasks, and 4) our proposed FedAL framework.

## RELATED WORK

Federated Learning

FL, an advanced distributed learning paradigm enabling multiple institutions to collaboratively train a model without data sharing (McMahan et al.), has opened up a new research field. The Federated Averaging (FedAvg) algorithm (McMahan et al.), a model-agnostic FL framework that aggregates the client models by calculating the weighted average value of model weights, has performed well in many fields. FedProx (Li et al.) is robust and has high performance in highly heterogeneous settings. VerifyNet (Xu et al.), the first verifiable FL framework, introduces a double-masking protocol to further guarantee the confidentiality of private data. FML (Shen et al.) allows more flexible choices of local models by training a generalized model collaboratively and a customized model independently. Although initially developed for edge devices (Kairouz et al.), FL has attracted increasing attention in healthcare due to its advantages of protection of patient data (Cheng et al.). FL was successfully applied to multi-institutional COVID-19 diagnosis (Bai et al.). Due to its task-agnostic characteristics, there has been considerable interest in integrating FL with other machine-learning paradigms. Yang et al. (Yang et al.) proposed a federated semi-supervised learning framework to thoroughly use unlabeled data. Dong et al. (Dong and Voiculescu) introduced contrastive learning to FL to make full use of decentralized unlabeled data. Liu et al. (Liu et al., 2021) proposed a problem setting called federated domain generalization to learn a federated model able to generalize to unseen target domains directly.

Active Learning

AL represents a significant field in human-in-the-loop machine learning (Budd et al.). It interactively queries valuable and adequate data to annotate for optimal model training, making it possible to achieve comparable performance with a model trained with full data while reducing the amount of training data. AL encompasses a variety of strategies (Budd et al.), such as uncertainty-based methods (Shi et al., Wen et al., Zhou et al.), which prioritize data presumed to contain more information due to their higher uncertainty. Shi et al.'s selective annotation framework and Zhou et al.'s Active, Incremental Fine-Tuning method are examples of this approach. Furthermore, representativeness-based methods, such as the conditional GAN-based method (Ravanbakhsh et al.), aim to select data samples that are indicative of the larger dataset. Advances in AL, including classifier-discrepancy methods (Cho et al.). and reinforcement learning techniques (Woodward and Finn), demonstrate the field's ongoing evolution and its increasing relevance to efficient model training.

## METHOD

This section provides details of our FedAL framework with an annotation strategy. An overview of our framework is shown in Figure 1.

Problem Formulation

As shown in FedAL framework in Table 1., initially, each *m*-th client contains an unlabeled (*U*) dataset $D_U^m$ and labeled (*L*) dataset $D_L^m$, which is initialized with randomly selected data. Our framework consists of two fundamental steps: *a)* AL and *b)* FL. The AL step is executed periodically before the execution of the FL step. The two steps are executed repeatedly until a satisfactory performance is achieved.

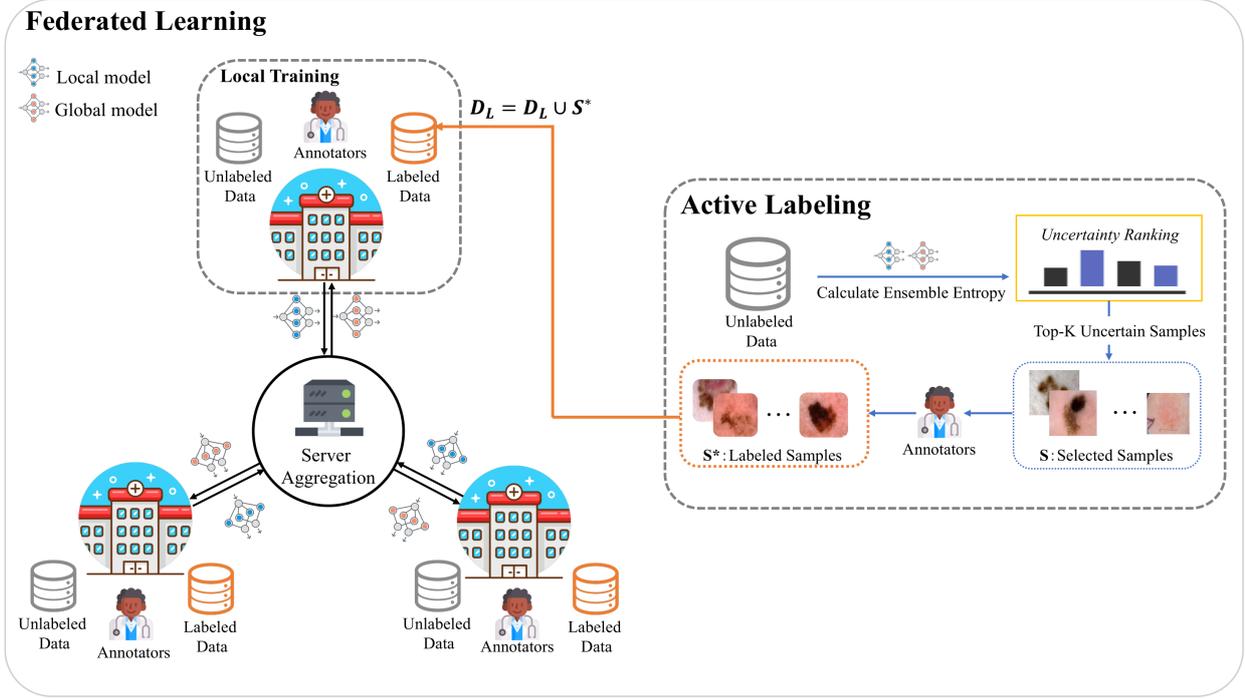

Figure 1. Overview of our FedAL framework for skin-lesion classification

Federated Active Learning

*AL Step*: The key idea of AL lies in the design of the sampling function. To make it more generalized, we use $A(\cdot)$ to denote the sampling function in the AL step, which is used to select the best data for models to train. In our FedAL framework, $A(\cdot)$ is an ensemble entropy (Beluch et al.) based sampling function, which is described in Section III.*C*. the $A(\cdot)$ can also be set to random sampling or sampling functions in AIFT (Zhou et al.), SA (Shi et al.), MCDAL (Cho et al.) or other AL methods. As the first federated active learning framework in natural images (Ahn et al.), the amount of sampled data for annotation in each client is proportional to the total amount of data with that client. This method accounts for the varying sizes of data across clients, ensuring a fair and efficient contribution of each client. Specifically, $a^m$, defined as the number of selected unlabeled data samples of the *m*-th client at *each* AL step, is given by

$$a^m = \frac{U^m \times \gamma}{Total\ AL\ rounds} \quad (1)$$

where $U^m$ is the total number of unlabeled data samples in the *m*-th client, $\gamma$ is the sample ratio of the data selected for annotation to the total amount of data, and *K* is the number of AL rounds.

The data selected by the sampling function in the *m*-th client at each AL step is $S_m = A(D_U^m, a^m)$. Those selected samples are then removed from the unlabeled dataset, annotated by human experts as $D_L^{m*} = annotate(S_m)$, then used to update labeled data $D_L^m \leftarrow D_L^m \cup D_L^{m*}$.

*FL Step*: With the active-sampling method, the potentially most valuable samples are annotated and added to the labeled dataset incrementally, which enables each client to fine-tune their model progressively under FL. In this study, FedAvg (McMahan et al.) was used in the FL step to aggregate client models.

Table 1.    FedAL framework

---

**Algorithm 1** FedAL

---

**Input**:

$M$: number of clients, $\{D_U^m\}_{m=1}^M$: unlabeled datasets, $\{D_L^m\}_{m=1}^M$: labeled datasets, $K$: AL-step interval rounds, $T$: number of FL rounds, $\psi_0$: initialized global model, $a^m$: amount of selected unlabeled data in AL step, $\eta_m$: learning rate for $m$-th client

**Output**:

$\{D_L^m\}_{m=1}^M$: labeled datasets, $\psi_T$ : optimal global model

**Functions**:

$A(\cdot)$: Sampling function, $L(\cdot)$: Loss function

**procedure RunServer()**

   **for** round $t = 1$ **to** $T$ **do**

      **for** client $m = 1$ **to** $M$ **in parallel do**

         $\phi_m^t \leftarrow \text{RunClient}(\psi_{t-1})$

      **end for**

$$\psi_t \leftarrow \sum_{m=1}^{M} \frac{n_m}{n} \phi_m^t$$

   **end for**

**end procedure**

**function RunClient($\phi$)**

    **if** $t \% K == 0$:

   **AL step**:

       selected samples $S_m^k = A(D_U^m, a^{m,k})$

       $D_L^{m*} = annotate(S_m^k)$

       $D_L^m \leftarrow D_L^m \cup D_L^{m*}$

       $D_U^m \leftarrow D_U^m - S_m^k$

   **Training step**:

       **for** epoch $e = 1$ to $E$ **do**

          **for** minibatch $d \in D_L^m$ **do**

             $\phi \leftarrow \phi - \eta_m \nabla L(\phi)$

          **end for**

       **end for**

     **return** $\phi$  to server

**end function**

---

   FedAvg contains three elemental steps. 1) The server sends the initial model $\psi$ to each client. 2) Each client trains its model $\phi_m$ locally and in parallel then sends the model weights (not patient data) to the server. 3) The server aggregates the collected client models $\{\phi_m\}_{m=1}^{M}$ to a global model $\psi$, which is used to replace all client models.

Step 2) and 3) are repeated for certain times. In FedAvg, the aggregation step is defined as

$$\psi \leftarrow \sum_{m=1}^{M} \frac{n_m}{n} \phi_m, \qquad (2)$$

where $M$ refers to the number of clients, $n$ is the total amount of training data in all clients, and $n_m$ denotes the number of training samples in the $m$-th client, $\phi_m$ is the local model of the $m$-th client, and $\psi$ is the global aggregated model.

Ensemble Entropy

The entropy-based method was selected for our FedAL framework due to its ability to offer a thorough analysis of uncertainty, simultaneously maintaining model-agnostic and hyperparameter-free characteristics of our method, enabling seamless integration with various FL frameworks. This approach guarantees the extraction of highly informative samples from both local and global models, thereby enhancing the overall performance of the federated system, and also provides the necessary flexibility to adapt our framework to various domains and tasks. We will further analyze and discuss the underlying reasons for this methodological choice in the discussion section of this paper.

Originating from information entropy (Shannon) in Shannon's information theory, the most common implementation in uncertainty-based AL is to choose data with the predicated probability distribution having the highest entropy (Shannon). Such data are believed to contain more information for training models. Ensemble-based AL methods have been proven to be powerful in alleviating labeling efforts (Beluch et al.). Inspired by that study, we use ensemble entropy as our selection criterion in the sampling function $A(\cdot)$, which is defined as

$$H[y|x, D_{train}] = -\sum_c \left(\frac{1}{T}\sum_t p(y=c|x, \phi_t)\right) \cdot \log\left(\frac{1}{T}\sum_t p(y=c|x, \phi_t)\right), \qquad (3)$$

where $T$ is the number of models in the ensemble, $\phi^t$ is the $t$-th model in the ensemble, and $p(y=c|x, \phi^t)$ is the predicted probability of $\phi^t$ for a given class $c$ and input $x$.

As indicated in the above study (Beluch et al.), ensemble methods have a potential drawback: they need to train multiple models to construct an ensemble, which can be costly in computation. Instead, clients in FL have local models and a global aggregated model during training, which means we do not need to train extra models to construct an ensemble. More specifically, the sampling function in FedAL is defined as

$$S_m^k = \underset{S \subseteq D_U^m,\ |S|=a^{m,k},\ x \in D_U^m}{\arg\max} H[y|x, D_{train}], \qquad (4)$$

where the ensemble in $H[y|x, D_{train}]$ for the $m$-th client is $\{\phi_m, \psi\}$, and $\psi$ is acquired from the server. In our experiments, $\phi$ is the model trained after the last round in the FL step.

Calculating entropy using both local and global models makes it feasible to select valuable data for both models, which can improve the performance of FL system. By applying sampling function (4) to each client in the AL step, we select samples having the top-k largest ensemble entropy in all unlabeled data. Those selected samples have the largest uncertainty for both local and global models.

## EXPERIMENTS

Real-world Skin Cancer FL Dataset

Skin cancer is one of the most common cancers worldwide. An accurate diagnosis method of skin lesions is essential to cure melanoma (Esteva et al.), which is known as the most severe type of skin cancer. We constructed a real-world skin cancer FL dataset in the same manner as in a previous study (Chen et al.), which was very heterogeneous, as shown in Table 2. Skin images of different clients were taken from different places, which means distributions among clients were non-IID. Specifically, 8490 and 2000 dermoscopic images were collected from two public dermoscopic datasets HAM10K (Tschandl et al.) and MSK (Codella et al.), respectively. With the location information of images provided by HAM10K (Tschandl et al.), we assigned the images generated in the same place into one client/hospital. Hence, the dataset included four hospitals, one hospital from MSK (Codella et al.) and three hospitals from HAM10K (Tschandl et al.). The detailed information on the dataset is summarized in Table 2. We further randomly divided the data from each hospital into training, validation, and test sets in the ratio 7:1:2.

Table 2. DETAILED INFORMATION ON FL DATASETS

| Hospital | Dataset | Skin-lesion Type | | | |
|---|---|---|---|---|---|
| | | Nevus | Benign Keratosis | Melanoma | Total |
| A | HAM10K (Tschandl et al.) | 803 | 490 | 342 | 1635 |
| B | HAM10K (Tschandl et al.) | 1832 | 475 | 680 | 2987 |
| C | HAM10K (Tschandl et al.) | 3720 | 124 | 24 | 3868 |
| D | MSK (Codella et al.) | 1372 | 254 | 374 | 2000 |

Implementations

We used ResNet-101 (He et al.) pre-trained with ImageNet (Russakovsky et al.) as the backbone network. In the local training at hospitals, the models were optimized using cross-entropy loss with the Adam optimizer (Kingma and Ba), where the learning rate was initialized as $2 \times 10^{-4}$, weight decay was set to $5 \times 10^{-4}$, and coefficients used for computing the running average of gradient and its square were 0.9 and 0.999, respectively. The total number of training epochs was 125. Communication in the FL step was conducted every five epochs. The AL step was conducted every 10 epochs until reaching 100 epochs. The last 25 epochs where no additional labeled data were added were used to fine-tune the model. Images were cropped with the central square regions using the shorter side length and resized to 224×224 then normalized to match the input size of the pre-trained ResNet-101 (He et al.) model. During training, images were augmented by a combination of random flip, rotation, translation, scaling and one of the blur transformations in Gaussian blur, Gaussian noise, and median blur. All experiments were conducted on NVIDIA A100 GPU.

Evaluation Metrics

In our study, we evaluated the performance of various FL methods in a multi-class setting using several metrics. Firstly, we utilized the validation sets to identify the most effective model based on predefined criteria such as Macro-F1, Micro-F1, and AUC. Once the optimal model was determined through validation, we then conducted tests on an independently held-out test set on each client. Macro-F1 measures the average performance of the classifier for each label, without taking the label distribution into account. Micro-F1 calculates the performance by aggregating the total true positives, false negatives, and false positives across all classes. Area Under the Receiver Operating Characteristic (AUC) Curve gauges model's ability to distinguish among multiple classes. In addition to showing the performance of the global model on each client, we also used the macro average of those metrics among all clients to compare the various methods. This approach ensures that each client contributes equally to the final result, regardless of the characteristics of their individual datasets, thereby providing a fairer comparison.

# EVALUATION

Performance Comparison

The comparison frameworks we investigated are: 1) FedAvg with random sampling, which is the lower bound of this task, 2) FedAvg with full-data training, which is the upper bound, 3) FedAvg with client-level AL, including AIFT (Zhou et al.) and SA (Shi et al.), which is claimed to be a state-of-the-art AL method in two skin-analysis tasks, and 4) a FedAL framework (Ahn et al.) using FedAvg with MCDAL (Cho et al.), which is claimed to be the state-of-the-art in several natural image classification tasks.

Table 3.  OVERALL QUANTITATIVE COMPARISONS

| Methods | Data Amount | Micro-F1 | Macro-F1 | AUC |
|---|---|---|---|---|
| **Centralized learning upper bound** | 100% | 87.20±0.35 | 79.62±0.26 | 92.07±0.44 |
| FedAvg + full-data (**upper bound**) | 100% | 86.31±0.98 | 78.53±0.30 | 91.89±0.33 |
| FedAvg + AIFT (Zhou et al.) | 50% | 83.89±0.80∗ | 75.11±1.58∗ | 89.68±0.83∗ |
| FedAvg + SA (Shi et al.) | 50% | 84.33±1.07∗ | 76.39±1.45∗ | 90.36±0.90 |
| FedAvg + MCDAL (Ahn et al., Cho et al.) | 50% | 82.12±1.16∗ | 74.14±1.38∗ | 89.91±0.75∗ |
| FedAL (ours) | 50% | **86.17±0.62** | **78.50±1.08** | **91.43±0.47** |
| FedAvg + random (**lower bound**) | 50% | 83.58±0.75∗ | 76.14±1.77∗ | 90.94±1.12 |

∗ indicates statistical significance against our FedAL

In our FedAL framework, on the basis of initial randomly selected labeled data, we continuously added training samples until the model attained performance comparable to the full data training. Using only up to 50% of the training data, the performance of our framework in terms of Micro-F1, Macro-F1, and AUC was 99.9, 100, and 99.4% of the upper bound (full-data training), respectively. The performance of our framework was comparable to that of full-data training, demonstrating its efficacy. The performance of our framework compared with random sampling in terms of Micro-F1, Macro-F1, and AUC was 3.11%, 3.15%, and 0.5%, respectively. All the results were averaged over 5 runs using different random seeds.

Table 4.  QUANTITATIVE COMPARISONS AMONG PERFORMANCE OF LOCAL DATASETS

| Methods | Data Amount | Macro-F1 | | | | |
| | | A | B | C | D | Avg |
|---|---|---|---|---|---|---|
| FedAvg + full-data (**upper bound**) | 100% | 78.3 | 82.7 | 82.9 | 70.1 | 78.5 |
| FedAvg + AIFT (Zhou et al.) | 50% | 73.4 | 78.4 | 81.6 | 67.1 | 75.1 |
| FedAvg + SA (Shi et al.) | 50% | 75.2 | 78.7 | 82.1 | 69.8 | 76.4 |
| FedAvg + MCDAL (Ahn et al., Cho et al.) | 50% | 73.8 | 75.4 | 79.5 | 67.7 | 74.1 |
| FedAL (ours) | 50% | **76.3** | **84.3** | **82.7** | **70.5** | **78.5** |
| FedAvg + random (**FL lower bound**) | 50% | 76.0 | 79.4 | 80.8 | 68.0 | 76.1 |
| **Localized learning lower bound** | 100% | 73.6 | 71.7 | 68.5 | 52.3 | 66.5 |

As shown in Table 2, our FedAL framework outperformed all other federated active learning models in terms of overall performance, and it showed statistically significant improvements over the competing methods for both Micro-F1 and Macro-F1 metrics (paired t-test: P < 0.05). Meanwhile, the differences in performance between our FedAL framework and the upper bound of full-data training were not statistically significant, as indicated by the paired t-test results (P = 0.868 for Micro-F1, P = 0.939 for Macro-F1, and P = 0.113 for AUC). These results indicate that our FedAL method attained a performance level on par with the full-data training benchmark, the latter being considered the optimal scenario in this context. In addition, we included the centralized training results as a reference, where we combined all client data to form a single training dataset. While Federated Learning (FL) is generally considered in the scenarios where direct data sharing between clients is not possible, there is still a trade-off between privacy and model performance in FL. Our FedAL approach achieved the results close to those of centralized training, as shown in Table 3. This similarity in performance, especially considering the reduced data requirements of our method, highlights its potential efficacy in practical applications. However, due to the relatively small performance gap in AUC between the lower and upper bounds, our method did not demonstrate a statistically significant improvement in AUC compared to some methods.

ResNet-101 (He et al.) was used as the backbone network for all frameworks compared in our experiments to remove the variations caused by different networks. As shown in Table 3., our framework outperformed all others. The performance of our framework was higher compared with the others. Table 4. shows that our framework also outperformed most frameworks in each local hospital. As in a recent FL study for COVID-19 where the global model trained by FL improved the performance of the locally trained models by 27.6% in AUROC (Soltan et al., 2024), FedAL method notably improved local models by an average of 18.0% in Macro-F1 score. This significant boost in performance underscores the efficacy of FedAL in the healthcare domain, particularly in enhancing performance and reducing annotation cost while preserve patient privacy.

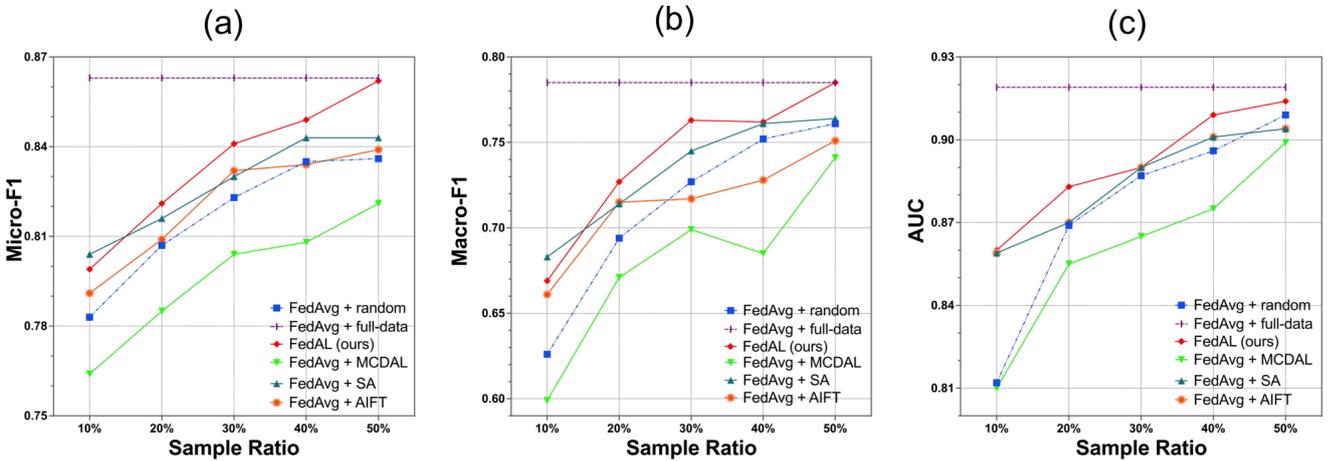

Figure 2. Performance comparison of global FL models against sample ratios: (a) Micro-F1, (b) Macro-F1, and (c) AUC

Figure 2 illustrates the performance of the frameworks with different sample ratios. As the sample ratio increased, the performance of our framework surpassed nearly all others and achieved a performance comparable to full-data training when the sample ratio was 50%. FedAvg with MCDAL performed worse than random selection. The reason might be that MCDAL is the only AL method that does not take uncertainty-based criterion into the sampling function.

Our framework achieved a marked increase in performance compared with random selection and achieved comparable results with full-data training, which indicates that it can select the most valuable, informative samples to label. Hence, it can significantly decrease the annotation workload in medical-image analysis, where annotation cost is extremely high.

FedAvg with AIFT (Zhou et al.) and FedAvg with SA (Shi et al.) performed marginally better than random selection and could not achieve satisfactory performance compared with full-data training. FedAvg with MCDAL (Ahn et al., Cho et al.) performed worse than random selection in the non-IID skin cancer task, while as claimed in a previous study (Ahn et al.), it achieved state-of-the-art performance in FedAL on several natural-image-classification tasks.

Ablation Study

To evaluate the contribution of the local and global models in evaluating the unlabeled data, we conducted experiments in which we only used the local model or global model to calculate the entropy at different sample ratios. As shown in Figure 3., with the sample ratio increase after 20%, using both models in our framework performed better than only the local model or global model. Loosely speaking, using only the local model will obtain a better Micro-F1 score, while using only the global model will obtain a better Macro-F1. Therefore, our framework has the combined advantages of both models to calculate ensemble entropy.

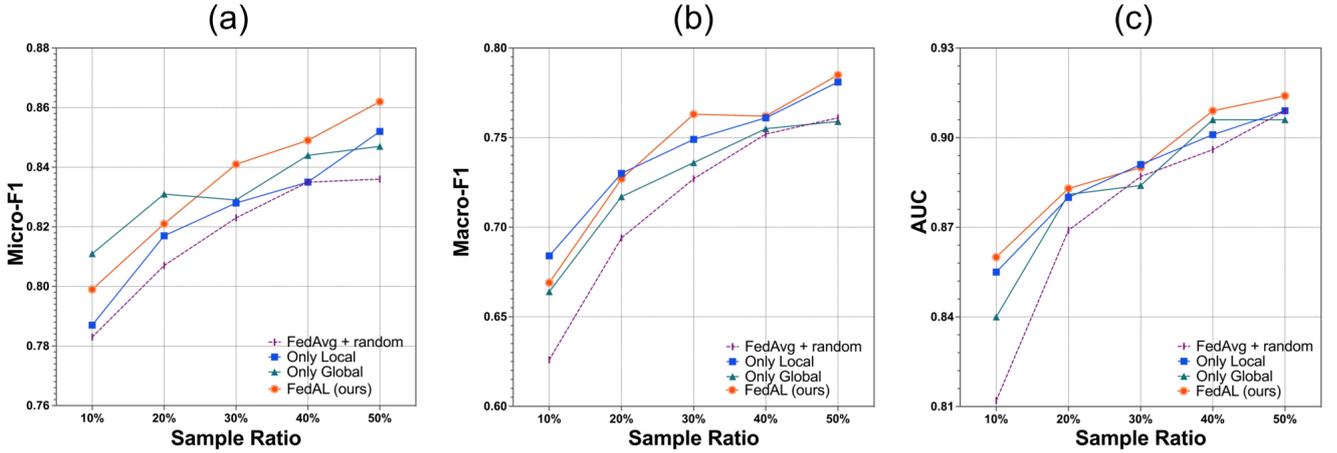

Figure 3.   Results of ablation study: (a) Micro-F1, (b) Macro-F1, and (c) AUC

## DISCUSSION

Analysis of Federated Active Learning

By introducing the AL step into FL, our FedAL framework has the advantages of 1) being able to train models collaboratively without sharing private data and 2) decreasing the annotation workload in medical-image analysis. As a pluggable module, our AL strategy is convenient because of its hyperparameters-free and model-agnostic characteristics. The extra time introduced by the AL step is trivial because we use current models to construct ensembles, and ensemble entropy is effortless to calculate.

Effectiveness of global models

In FL, the local and global models can be mutually affected, which is essential to achieving satisfactory performance collectively. Although our FedAL framework and federated learning frameworks with client-level AL methods, i.e., FedAvg with SA (Shi et al.) and FedAvg with AIFT (Zhou et al.), are all uncertainty-based, client-level frameworks, only take into account the uncertainty of the local model. On the contrary, uncertainty estimation in our framework takes both local and global models into account, enabling it to select the samples that are valuable and informative for both models and perform better. In conclusion, we believe that introducing both local and global models to design the sampling function can improve the performance of FL models under AL scenario.

Hyperparameters in Active Learning

Most AL studies used fully annotated datasets to simulate the annotation process, where labels can be acquired immediately after the AL-query step. However, in real-life medical-image-analysis scenarios, it takes considerable time to receive feedback or annotations after the execution of the AL-query step. The selection of data to be labeled is also an irreversible process. The same hyperparameters suitable in one specific experimental dataset usually cannot be extended to other datasets. Therefore, manual hyperparameter adjustment cannot be ideally applied in most real-life scenarios.

In addition to superior performance, our framework has no hyperparameters in the AL step, while both FedAvg with SA (Zhou et al.) and AIFT (Zhou et al.) use hyperparameters to adjust the trade-off between informativeness and representativeness criteria in sample selection, making them difficult to be deployed in non-simulated scenarios. In a real-life AL scenario, we have no chance to adjust the hyperparameters of our framework in sample selection. Therefore, it is much more robust and reliable than other federated active learning frameworks containing hyperparameter adjustments.

<u>Extensibility</u>

Most AL methods are designed to address specific modalities and tasks. AIFT (Zhou et al.) uses multiple patches of an image to evaluate informativeness. SA (Shi et al.) uses Principal Component Analysis (PCA) features of images to do clustering. Therefore, AIFT and SA can only be applied to image-classification tasks due to their special processing in the image domain. The AL step in our framework can be directly extended to other modalities since our framework only relies on the probability predicted by the model. It is also convenient to extend FedAL to recognition, segmentation, or other tasks in which the model used can output probability that can then be used to calculate the ensemble entropy to evaluate the uncertainty. The model-agnostic nature of our framework also enables us to incorporate other efficient models designed for limited training data, such as those developed by Suzuki et al. (Suzuki et al., 2003, Tajbakhsh and Suzuki, 2017). By merging these techniques, we aspire to further enhance the versatility and applicability of our FedAL framework.

<u>Limitations and Future Directions</u>

In our study, we acknowledge several limitations that highlight areas for future research. First, our methodology's proportional sampling approach, while ensuring equitable representation across diverse datasets, may underrepresent unique or critical information on minority classes, particularly in 'long-tailed' datasets (Shang et al., 2022) or those with unique classes (Zhang et al., 2023). This could affect our model's ability to capture the full range of data variability, indicating a need for alternative sampling strategies or enhanced representation of smaller datasets on clients in FL scenario or minority classes globally. Additionally, the publicly available datasets we used in this study lack Fitzpatrick skin type annotations, which limits the scope of our analysis across different skin types and potentially affects the depth and diversity of these analyses. To address this limitation, we intend to apply our method to various datasets that include various types of skin lesions, such as the DDI dataset (Daneshjou et al., 2022), in future research. Lastly, our study's focus on data annotation efficiency within the federated learning context led to less exploration of special scenarios such as data centers with noisy labels (Yang et al., 2022), noisy data (Tuor et al., 2021), missing disease categories (Zhang et al., 2023), unbalanced data (Wang et al., 2021), among other unique challenges. Addressing these limitations in future work could further enhance the generalizability and applicability of our model. Furthermore, the hyperparameters-free and model-agnostic nature of our approach allows easy applications in other challenging scenarios or integration with different models. This versatility highlights our method's capacity to contribute broadly to federated learning and other relevant fields.

## CONCLUSION

We presented a novel FedAL framework for efficient labeling in skin-lesion analysis that performed better than other frameworks, which is hyperparameter-free and task-agnostic. The "hyperparameter-free" signifies that FedAL is ready to use upon implementation, eliminating the need for medical professionals to engage in complex setup procedures. The "task-agnostic" quality of FedAL denotes its broad applicability across different diagnostic objectives, such as segmentation, which illustrates its capacity to adapt to various diagnostic requirements in dermatology. In an era where patient confidentiality is paramount, our approach allows dermatologists to benefit from AI advancements in skin lesion analysis without the need to share sensitive patient data outside their hospitals. Moreover, FedAL integrates with the clinical workflow by involving the medical experts in the loop of data annotation within federated learning. This innovative method cuts down the time when doctors spend on annotation by up to 50%, which streamlines their workload while maintaining high accuracy in the AI's performance. It meets the need of the annotation strategy in FL in medical-image analysis. Specifically, we use ensemble entropy to evaluate the uncertainty of data then annotate the samples with the highest uncertainty in clients, where the ensemble is composed of the global and local models in each client. Extensive experiments demonstrated that our framework was able to achieve state-of-the-art performance on a real word dermoscopic FL dataset.

ACKNOWLEDGEMENTS

The authors are grateful to Dr. Ze Jin for his valuable suggestions and the members of the Suzuki Lab and Jinkun You from University of Macau for their valuable discussions. This research was supported by JST-Mirai Program Grant Number JPMJMI20B8, Japan.